%% file: main.tex
\documentclass[10pt,twocolumn,letterpaper]{article}

\usepackage{listings}
\usepackage[linesnumbered,ruled,vlined]{algorithm2e}

\usepackage[utf8]{inputenc}

\usepackage{iccv}              


\definecolor{iccvblue}{rgb}{0.21,0.49,0.74}
\usepackage[pagebackref,breaklinks,colorlinks,allcolors=iccvblue]{hyperref}

\def\paperID{17} 
\def\confName{ICCV}
\def\confYear{2025}

\title{HA-RDet: Hybrid Anchor Rotation Detector for Oriented Object Detection}

\author{Phuc Nguyen\\
University of Information Technology, Vietnam\\
{\tt\small phucnda@gmail.com / 20520276@gm.uit.edu.vn}
}

\begin{document}
\maketitle
\input{sec/0_abstract}    
\input{sec/1_intro}
\input{sec/2_relatedwork}
\input{sec/3_methodology}
\input{sec/4_experiments}

\input{sec/5_conclusion}
{
    \small
    \bibliographystyle{ieeenat_fullname}
    \bibliography{main}
}

\end{document}

%% file: sec/0_abstract.tex
\begin{abstract}
Oriented object detection in aerial images poses a significant challenge due to their varying sizes and orientations. Current state-of-the-art detectors typically rely on either two-stage or one-stage approaches, often employing Anchor-based strategies, which can result in computationally expensive operations due to the redundant number of generated anchors during training. In contrast, Anchor-free mechanisms offer faster processing but suffer from a reduction in the number of training samples, potentially impacting detection accuracy. To address these limitations, we propose the Hybrid-Anchor Rotation Detector (HA-RDet), which combines the advantages of both anchor-based and anchor-free schemes for oriented object detection. By utilizing only one preset anchor for each location on the feature maps and refining these anchors with our Orientation-Aware Convolution technique, HA-RDet achieves competitive accuracies, including 75.41 mAP on DOTA-v1, 65.3 mAP on DIOR-R, and 90.2 mAP on HRSC2016, against current anchor-based state-of-the-art methods, while significantly reducing computational resources. Source code: \url{https://github.com/PhucNDA/HA-RDet}
\end{abstract}

%% file: sec/1_intro.tex
\section{Introduction}
\label{sec:intro}

Object detection in aerial images (ODAI) stands as a pivotal challenge within the domains of computer vision and pattern recognition, boasting numerous practical applications \cite{yolo, yolov2, fasterrcnn, fastrcnn, reppoint, retinanet}. Furthermore, there exists a pressing need within the research community to explore advanced technologies capable of autonomously analyzing vast-scale remote sensing images. However, unlike scenes captured from horizontal perspectives, aerial images obtained from an overhead view often feature densely populated objects appearing in various scales, shapes, and arbitrary orientations. Such complexities pose formidable challenges for models, especially those relying on Horizontal Bounding Boxes (HBB) to accurately represent object instances in the image. To address this, Oriented Bounding Boxes (OBB) are utilized to offer a more precise depiction of object locations by incorporating additional directional information. Notably, several significant challenges emerge in the realm of detecting objects in aerial images, including:

\begin{itemize}
  \item\textbf{Large aspect ratio}: Objects in aerial images often have irregular shapes, resulting in high aspect ratios, such as bridges, ships, harbors, etc.
  \item\textbf{Scale variations}: Varying ground sampling distance (GSD) of sensors leads to scale variation among objects captured by different sensors in the same scene \cite{adn_ood}.
  \item\textbf{Dense arrangement}: Aerial images usually contain multiple objects distributed densely, such as ships in a harbor or vehicles in a parking lot, which can pose a big challenge for object detection algorithms as they need to accurately identify and distinguish object instances.
  \item\textbf{Arbitrary orientations}: Objects in aerial images can appear in various orientations, requiring object detection models to have precise orientation estimation capabilities.
\end{itemize}

Existing oriented detectors \cite{roitrans, glidingvertex, redet, orientedrcnn, aood} typically rely on proposal-driven frameworks utilizing a Region Proposal Network (RPN), which employs numerous anchors with varying scales and aspect ratios to generate Regions of Interest (RoIs). However, horizontal RoIs often cover multiple instances, causing ambiguity, while rotated RoIs improve recall but incur high computational costs. Hence, developing an efficient region proposal network capable of generating high-quality proposals is crucial to overcome accuracy and computational limitations in current state-of-the-art oriented detectors. We identify two key reasons for the lack of efficiency in region proposal-based oriented detectors: \textbf{(1)} Most oriented object detection methods adopt Anchor-based or Anchor-free training approaches, each with its own challenges and trade-offs in terms of performance and computational efficiency. Anchor-based methods achieve good performance but are sensitive to oriented anchor hyperparameters, while Anchor-free methods accelerate training and inference speed but suffer from decreased detection performance due to shortage number of preset training anchors. \textbf{(2)} The conventional convolution technique, designed with fixed receptive fields and aligned with axes, is ill-suited for objects with arbitrary orientations and varying scales in aerial images. Predefined anchor representations often fail to fully capture the shape and orientation of objects, leading to misalignment between fixed convolution features and object orientations, thus hindering effective object localization and classification in oriented object detection models.

This study presents a novel method, Hybrid-Anchor Rotation Detector (HA-RDet), for oriented object detection. HA-RDet combines the strengths of anchor-based and anchor-free paradigms through a hybrid anchor training scheme that employs a single anchor per location on the feature map. At the core of our design is the proposed Orientation-Aware Convolution (O-AwareConv), which serves as a bridge between one-stage and two-stage detection heads. O-AwareConv improves proposal quality by dynamically adapting to the object’s shape and orientation.
Extensive experiments on standard oriented object detection benchmarks demonstrate the effectiveness of our approach. HA-RDet addresses key challenges such as insufficient positive training samples and over-encoding of geometric information by initially reducing anchor hyperparameters and subsequently refining proposals in later stages.
Unlike prior methods, HA-RDet achieves a superior balance between accuracy and computational efficiency in both training and inference. It requires fewer resources while maintaining competitive performance, thereby accelerating the detection process. Overall, HA-RDet offers an elegant and efficient region proposal-based solution that bridges the gap between anchor-based and anchor-free approaches, providing a strong baseline for future research in oriented object detection.


Our contributions are summarized as follows: 
\begin{itemize}
    \item Presenting the Hybrid-Anchor Rotation Detector that innovatively combines anchor-based and anchor-free techniques for oriented object detection
    \item Introducing the novel O-AwareConv for enhanced orientation-sensitive feature extraction
    \item Extensive experiments are conducted on challenging oriented object detection datasets to demonstrate the effectiveness of our proposed method.

\end{itemize}

%% file: sec/2_relatedwork.tex
\section{Related works}
\label{sec:relatedwork}


With the surge of machine learning, particularly deep learning, object detection has witnessed significant advancements, dividing into two main categories: two-stage detectors and one-stage detectors. Two-stage detectors begin by generating a sparse set of Regions of Interest (RoIs) in the first stage, followed by RoI-wise bounding box regression and object classification in the second stage. In contrast, one-stage detectors directly detect objects without the need for RoI generation. Furthermore, two fundamental training strategies, Anchor-free and Anchor-based methods, are crucial for object detection. While Anchor-free methods excel in rapid object detection, they suffer from a scarcity of positive training samples, resulting in performance degradation. Conversely, Anchor-based methods generate numerous anchors, facilitating an abundant supply of positive training samples and achieving excellent performance, albeit at a higher computational cost. Our HA-RDet integrates both training strategies, utilizing only one preset anchor for each feature map's location, thereby significantly reducing resource requirements while maintaining asymptotically state-of-the-art performance.

\subsection{Anchor-Based Detectors}
Most methods in object detection rely on predefined generated anchors, where multiple preset anchors with varying scales and ratios are assigned to each specific location on a feature map. For instance, ROI Transformer generates preset horizontal anchors and then applies RRoI Leaner and RRoI Wrapping for robust feature extraction. Oriented R-CNN proposed by \citeauthor{orientedrcnn} \cite{orientedrcnn} adopts an anchor-based approach with an Oriented Region Proposal Network, along with the Midpoint-offset representation. This representation encodes horizontal anchors into six distinguishable learnable parameters, ensuring a reasonable transformation from horizontal generated anchors to oriented proposals. Similarly, Faster R-CNN \cite{fasterrcnn} introduces the Region Proposal Network with nine multi-scale heuristically tuned anchors for every location on a feature map. These densely generated anchors are assigned either positive or negative labels based on the Intersection over Union (IoU) with a ground-truth box. Additionally, SSD \cite{ssd} enhances detector efficiency by directly predicting object location offsets and corresponding categories. While anchor-based methods are well-studied and yield remarkable performance by generating a large number of anchors with different ratios and scales for each location, they require significant computational resources for training and inference, making them impractical for oriented object detection in aerial images datasets.
\subsection{Anchor-Free Detectors}
The anchor-free mechanism differs from anchor-based strategies by utilizing only one key point for each location on a feature map, implying that each point on the feature map corresponds to a single anchor with a specific scale and ratio. $S^2$A-Net \cite{$S^2$A-Net} introduced the single-shot alignment network featuring the Feature Alignment Module and Oriented Detection Module, which ensure feature alignment using the Alignment Convolution specifically designed for oriented feature extraction. YOLO \cite{yolo} divides the image into an $S\times S$ grid, with each grid cell responsible for determining the presence of an object's center. CornerNet \cite{cornernet} predicts the final bounding box using a pair of key points (top-left and bottom-right) and undergoes a complex post-processing phase to combine key points of the same object. FCOS \cite{fcos} assigns all locations within an object's ground truth as positive, incorporating a "centerness" branch to suppress low-quality bounding boxes, thereby enhancing overall performance. Despite being faster, anchor-free methods may experience a performance decline compared to previous anchor-based methods.

%% file: sec/3_methodology.tex
\section{Proposed Hybrid-Anchor Rotation Detector}


\begin{figure*}[t]
\centering\includegraphics[width=.7\linewidth]{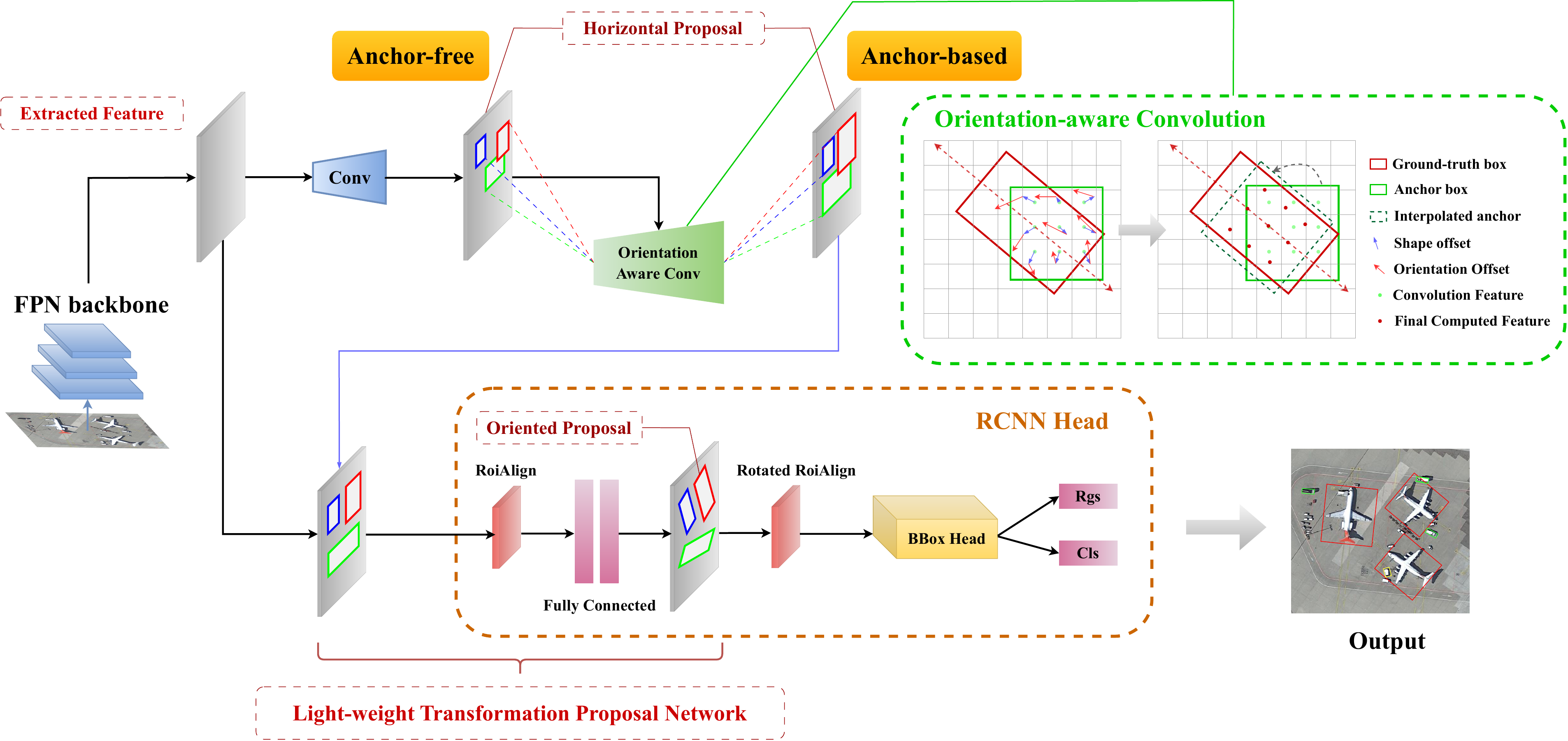}
\caption{Illustration of the architecture of the proposed HA-RDet. At first, an aerial image is passed through the FPN backbone to extract deep features. Then, the extracted features are fed through the Hybrid-Anchor RPN to produce horizontal anchors via an Anchor-free assigner strategy on rectangularized oriented ground truths. The anchors are refined with the novel Orientation-aware Convolution and the later Anchor-based head to produce high-quality horizontal proposals. After that, the lightweight proposal transformation network learns to produce oriented proposals from predicted RoI Align processed horizontal proposals. Finally, these oriented proposals are refined through oriented bounding box heads for classification and bounding box regression.}
\label{fig:model}
\end{figure*}

Most oriented object detectors follow the anchor-based strategy for predicting the 5-D encoded oriented bounding boxes. They often generate multiple anchors of different ratios and scales to allocate many instances of various shapes and sizes \textit{greedily}. This approach requires a large number of anchors, leading to high computational requirements. On the other hand, the anchor-free mechanism uses only a single anchor per location on a feature map, but it suffers from accuracy deterioration due to the lack of positive training samples. Many studies improve the performance of either method via post-refining, new assigner strategies, new anchor definitions, etc. Two strategies standstill and need to be fully exploited in oriented object detection.

In this section, we reason our proposals and explain how the strategy is applied and operated appropriately under different anchor encoder-decoder fashion.

\subsection{Hybrid-Anchor RPN} \label{section_hybrid_anchor_rpn}
\textbf{Anchor encoder-decoder strategy.} 
We begin by examining the vanilla baseline of HA-RDet, which builds on RetinaNet \cite{retinanet}—a detector originally developed for generic horizontal object detection—adapted here for oriented object detection by representing anchors in a 5D format $(x, y, w, h, \theta)$. The Anchor-free head, guided by an Anchor-free assigner, is trained to generate oriented region proposals directly from feature maps produced by standard 2D convolution (2DConv). In the next stage, the Anchor-based head refines these anchors using a 5D encoder-decoder to regress offsets $(\delta x, \delta y, \delta w, \delta h, \delta \theta)$, resulting in the final oriented object proposals. However, this experimental setup proves ineffective for two main reasons, both of which compromise proposal quality and consequently degrade the performance of the anchor-based head:
\textbf{(1)} \textit{Limited positive training samples results in insufficient and unreliable proposals from the anchor-free generator;}
\textbf{(2)} \textit{Overly complex five-dimensional anchor encoding in the anchor-free head introduces noise and reduces proposal precision.}

Our proposed approach (depicted in Fig. \ref{fig:model}) tackles these challenges by introducing the Hybrid-Anchor RPN, which simplifies anchor encoding from 5D $(x, y, w, h, \theta)$ to 4D $(x, y, w, h)$ while ensuring consistency between the two heads. We achieve this by utilizing the Anchor-free assigner strategy on rectangularized oriented ground truths for horizontal anchors, significantly increasing the number of positive samples. This augmentation aids the model in learning sufficient information for further refinement. Subsequently, the resulting horizontal anchors undergo processing in the Anchor-based stage, producing high-quality horizontal proposals for downstream tasks. This design guarantees an adequate number of positive samples for training and gradually enhances the quality of the final proposals. Detailed insights into the algorithmic implementation of the Hybrid-Anchor RPN are provided in Algorithm \ref{algorithm_rpn}.

\noindent{\textbf{Anchor-Free head.}} We use one preset anchor with a single scale and ratio for each location on a feature map. 
In the Anchor-free head, the normal RetinaNet \cite{retinanet} architecture used as a backbone encodes and decodes the horizontal anchors $(a_x,a_y,a_w,a_h)$ with the rectangularized ground truth target $(t_x,t_y,t_w,t_h)$ as the following Eq. \ref{equa:anchorgen1}:

\begin{equation}
    \begin{cases}
    \delta_x=(t_x-a_x)/a_w, & \delta_y=(t_y-a_y)/a_h\\
    \delta_w=\log(t_w/a_w),   & \delta_h=\log(t_h/a_h)\\
    a'_x=\widehat{\delta_x}a_w+a_x, & a'_y=\widehat{\delta_y}a_h+a_y\\
    a'_w=a'_w\exp(\widehat{\delta_w}),   & a'_h=a'_h\exp(\widehat{\delta_h})\\
    \end{cases}
    \label{equa:anchorgen1}
\end{equation}


where $\delta = (\delta_x,\delta_y, \delta_w, \delta_h)$ and $\widehat{\delta}=(\widehat{\delta_x},\widehat{\delta_y}, \widehat{\delta_w}, \widehat{\delta_h})$ are predicted encoded value and target encoded value, respectively. Having regressed $\widehat{\delta}$, we can easily decode the proposal $(a'_x,a'_y, a'_w, a'_h)$. With the classification branch omitted, the Anchor-free head aims to minimize the bounding box loss function $L_{AF}$:

\begin{equation}
    L_{AF}(\delta,\widehat{\delta})= w^{AF} [-\ln(\frac{Intersection(\delta, \widehat{\delta})}{Union(\delta, \widehat{\delta})})]
    \label{equa:anchorgen2}
\end{equation}

where $w^{AF}$ is the loss weight used to fine-tune the model. At the end of this stage, from a limited number of preset anchors, we obtain a preliminary set of horizontal candidates, prepared for further refinement.
\\

\begin{figure*}[t]
  \centering
   \includegraphics[width=0.68\linewidth]{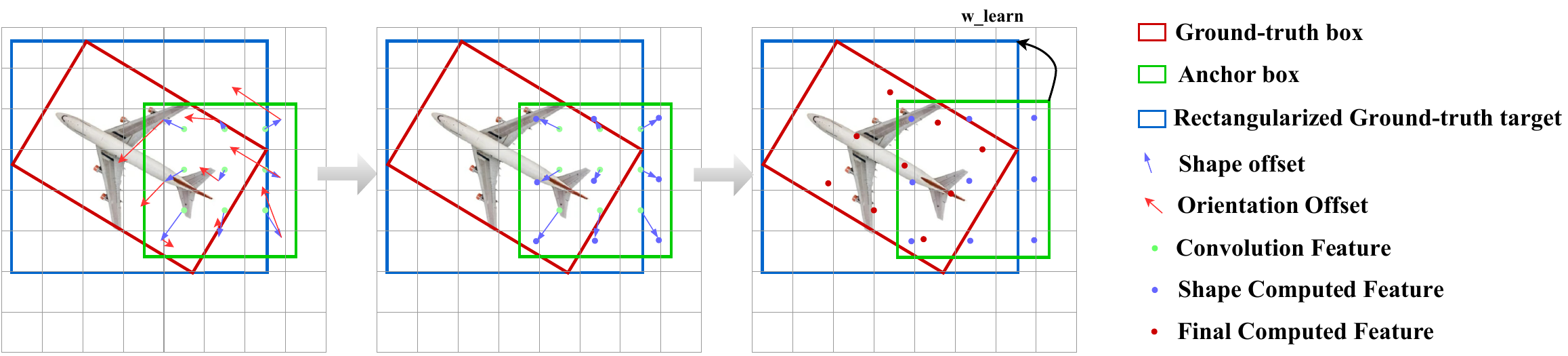}
   \caption{The shape offsets are computed via anchor size; ground-truth orientation is adopted to perform orientation offsets calculation. Shape offsets and orientation offsets produce the final features used for training the Anchor-based refinement stage. The final computed features are orientation-awareness, surging HA-RDet performance significantly.}
\label{fig:oawareconv}
\end{figure*}

\noindent{\textbf{Orientation-aware Convolution.}} To extract meaningful geometric cues from the set of predicted candidates, we propose orientation-aware convolution (O-AwareConv), which improves the alignment of global features and allows the extraction of orientation-sensitive features for horizontal anchor candidates. O-AwareConv adaptively incorporates the orientation of the ground truth object $\theta_{GT}$ during training. Specifically, it calculates a shape offset field for each horizontal anchor, in combination with an orientation offset derived from the rotation matrix $R(\theta_{GT})^T$, where $T$ denotes the transposition of the matrix.
These offsets guide the convolution to adjust its sampling locations, enabling it to align with the true orientation and geometry of objects. As illustrated in Fig.~\ref{fig:oawareconv}, this design transforms conventional convolutional features into orientation-aware representations, tailored to each anchor-ground-truth pair.

Formally, for each spatial location $p$ on a feature map $X$, O-AwareConv uses the anchor candidate $a^* = (x, y, w, h)$, stride $S$, kernel weights $W$, and kernel size $k$ to calculate the position $\Omega$ of the final compute feature $Y$ as defined in Eq.~\ref{equa:horizontalalign2}:

\begin{equation}
    \begin{cases}
    Y(p)=\sum_{r\in Kernels ; o \in \Omega} W(r)\cdot X(p+r+o) \vspace{0.75em} \\ 
    \Omega=\frac{1}{S}(\textit{\textbf{(x,y)}}+\frac{1}{k}\textit{\textbf{(w,h)}}\cdot r \cdot R({\theta_{GT}})^{T})-p-r \vspace{0.75em} \\
    R(\alpha)=\begin{bmatrix}
    cos{\alpha} & -sin {\alpha} \\
    sin {\alpha} & cos {\alpha}
    \end{bmatrix}
    \end{cases}
    \label{equa:horizontalalign2}
\end{equation}

In our approach, each anchor is assigned to at most one ground-truth object, while each ground truth can correspond to multiple anchors. To make full use of the training data, we retain all anchor candidates, including negative ones, in the previous anchor-free heads as a set of potential predefined candidates, anticipating that their quality will improve progressively during the refinement stage. During inference, when ground-truth annotations are not available, the orientation offset field is set to zero, ensuring deterministic feature offset computation.



\begin{figure*}
    \centering
    \includegraphics[width=10cm]{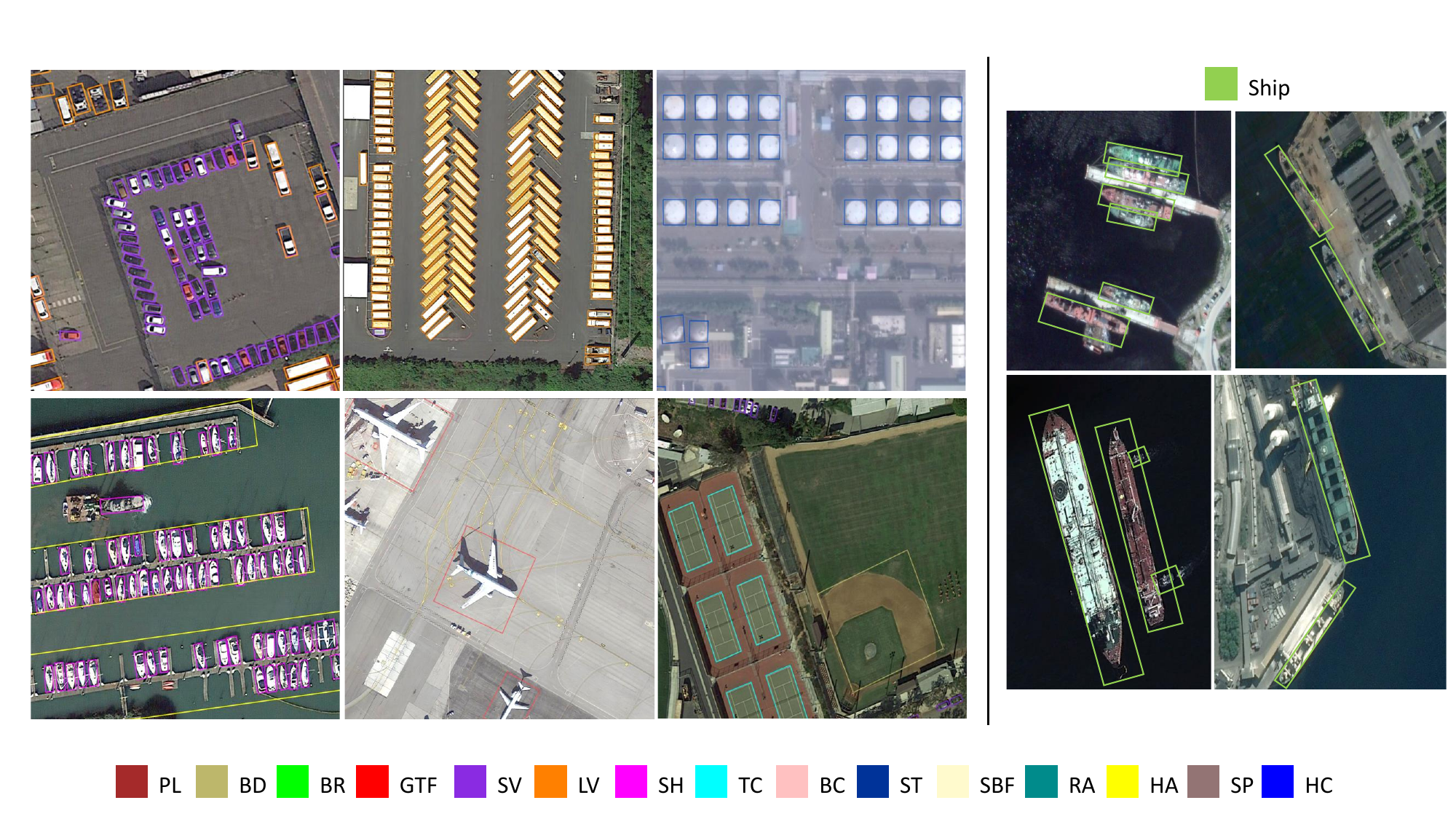}
    \caption{\textbf{Left: }Detection results on DOTA. \textbf{Right: }Detection results on HRSC2016} 
    \label{fig:resultdota}
\end{figure*}

\vspace{5mm} 

\lstset{escapeinside={<@}{@>}}
\begin{algorithm}[h]
    \caption{Hybrid-Anchor training strategy}
    \label{algorithm_rpn}
    \LinesNotNumbered
\fontsize{9}{11}\selectfont
\vspace{7.5px}

\KwIn{
\begin{itemize}
    \vspace{1.5px}
    \itemsep 0.25em
    \item \fontsize{7.5}{11}\selectfont\texttt{X: }
            \fontsize{9pt}{11}\selectfont{Multi-level features}
    \item \fontsize{7.5}{11}\selectfont\texttt{gt: }
            \fontsize{9pt}{11}\selectfont{Ground truth}
    \item \fontsize{7.5}{11}\selectfont\texttt{rgt: }
            \fontsize{9pt}{11}\selectfont{Rectangularized ground truth}
\end{itemize}
}
\vspace{7.5px}
\KwOut{
\begin{itemize}
    \vspace{1.5px}
    \itemsep 0.25em 
    \item \fontsize{9pt}{11}\selectfont{Region proposals}
\end{itemize}
}

\vspace{5.5px}
\texttt{\textcolor{red}{\textbf{Anchor-free stage}}}
\fontsize{7.5}{11}\selectfont
\begin{verbatim}
 1. A = anchor_generator(X)
 2. target = assigner(A, rgt)[`index']
 3. X = 2Dconv(X)
 4. bbox_pred = regression(X)
 5. A = decode(A, bbox_pred)
 6. loss_af = box_loss(A, rgt[target])
\end{verbatim}
    
    
    



    

\fontsize{9}{11}\selectfont
\texttt{\textcolor{red}{\textbf{Anchor-based stage}}}
\fontsize{7.5}{11}\selectfont

\begin{verbatim}
 7. target = assigner(A, rgt)[`index']
 8. offset = shape_offset(A)
 9. or_offset = or_offset(A, gt[target])
10. offset = dot(offset, or_offset)
11. X = OAconv(X, offset)
12. bbox_pred = regression(X)
13. bbox_cls = classification(X)
14. A = decode(A, bbox_pred)
15. loss_rgs = box_loss(A, rgt[target])
16. loss_cls = objectness(bbox_cls, label[target])
17. loss_ab = loss_rgs + loss_cls
\end{verbatim}

\fontsize{9}{11}\selectfont
\texttt{\textcolor{red}{\textbf{Return}}}
\fontsize{7.5}{11}\selectfont
\begin{verbatim}
19. loss_rpn = loss_af + loss_ab
20. proposals = refinement(A, bbox_cls)
\end{verbatim}

\vspace{5.5px}
\fontsize{9}{11}\selectfont{\textbf{\underline{APPENDIX}}}

\begin{itemize}
    \itemsep 0.25em 
    \item \fontsize{7.5}{11}\selectfont\texttt{target: }
            \fontsize{9pt}{11}\selectfont{Target ground-truth}
    \item \fontsize{7.5}{11}\selectfont\texttt{loss\_af: }
            \fontsize{9}{11}\selectfont{Anchor-free loss}
    \item \fontsize{7.5}{11}\selectfont\texttt{loss\_ab: }
            \fontsize{9}{11}\selectfont{Anchor-based loss}
    \item \fontsize{7.5}{11}\selectfont\texttt{decode(*): }
            \fontsize{9}{11}\selectfont{Bounding box decoder}
    \item \fontsize{7.5}{11}\selectfont\texttt{dot(*): }
            \fontsize{9}{11}\selectfont{Dot product mathematical operation}
    \item \fontsize{7.5}{11}\selectfont\texttt{assigner(*): }
            \fontsize{9}{11}\selectfont{Anchor assigner strategy}
    \item \fontsize{7.5}{11}\selectfont\texttt{OAconv(*): }
            \fontsize{9}{11}\selectfont{Orientation-aware Convolution}
    \item \fontsize{7.5}{11}\selectfont\texttt{shape\_offset(*): }
            \fontsize{9}{11}\selectfont{Get Shape offset}
    \item \fontsize{7.5}{11}\selectfont\texttt{or\_offset(*): }
            \fontsize{9}{11}\selectfont{Get Orientation offset}
    \item \fontsize{7.5}{11}\selectfont\texttt{refinement(*): }
            \fontsize{9}{11}\selectfont{Box refinement via NMS}
\end{itemize}
\end{algorithm}

Compared to standard 2D convolution, which assumes objects are axis-aligned, our method incorporates Deformable Convolution (DeformConv) \cite{dcn} with augmented offsets to improve spatial adaptability. However, DeformConv can still sample from suboptimal locations, especially under weak supervision and in densely populated scenes. To mitigate this, our O-AwareConv is used for feature extraction during training for each generated horizontal anchor, enabling robust orientation-aware alignment within the Hybrid-Anchor RPN.
Unlike 2DConv and DeformConv, the offset field in O-AwareConv is explicitly derived from the object’s shape and orientation. As shown in Fig.~\ref{fig:oawareconv}, this mechanism allows for more accurate and geometry-aligned feature extraction. During inference, O-AwareConv effectively captures global features without relying on orientation supervision, thereby improving anchor prediction accuracy.
At the end of this stage, we obtain a set of horizontal candidates from the Anchor-free head, along with their corresponding orientation-aware features $Y$ extracted by O-AwareConv.

\noindent{\textbf{Anchor-Based head.}}  The anchor-based head serves as a refinement module. It re-encodes the input according to Eq.~\ref{equa:anchorgen1}, taking as input the anchors generated in the previous stage along with their features extracted by the orientation-aware convolution. Its goal is to produce refined, high-quality proposals. The Anchor-based bounding box loss function $L_{AB}$ is derived from Eq. \ref{equa:anchorgen2}, albeit with a different weight $w^{AB}$. However, unlike Anchor-free, the classification branch with binary cross-entropy loss is used in this stage to estimate the objectness of the predicted region proposal. The overall loss of our Region Proposal Network is the sum of the Anchor-free loss $L_{AF}$ and the Anchor-based loss $L_{AB}$, denoted as $L_{RPN}=L_{AF}+L_{AB}$.
Subsequently, we obtain the final set of horizontal region proposals, which effectively capture orientation-sensitive objects and are ready to be transformed into oriented proposals.

\subsection{R-CNN oriented bounding box heads}

Having regressed well-qualified horizontal proposals from the Hybrid-Anchor RPN, we implemented a lightweight multi-layer neural network to transform the horizontal to oriented region proposals. RoIAlign \cite{maskrcnn} with bilinear interpolation is adopted to our baseline to extract the region of interest features. We then use 5D offset value representation $(\delta x, \delta y, \delta w, \delta h, \delta \theta)$ to encode the oriented proposals from horizontal ones and perform simple regression using two fully-connected layers.

For the oriented bounding box head, we adopt the Rotated RoIAlign operation \cite{orientedrcnn} to extract rotation-sensitivity features from each oriented proposal.

%% file: sec/4_experiments.tex
\section{Experiments}

\subsection{Datasets}

\textbf{DOTA-v1.0} \cite{dota} is a large-scale dataset comprising high-quality aerial images for detecting objects, containing $2,806$ high-defined images with resolutions $4000 \times4000$. It classifies annotated instances into $15$ common categories: plane (PL), baseball diamond (BD), bridge (BR), ground track field (GTF), small vehicle (SV), large vehicle (LV), ship (SH), tennis court (TC), basketball court (BC), storage tank (ST), soccer ball field (SBF), roundabout (RA), harbor (HA), swimming pool (SP) and helicopter (HC). During the training and testing process, the images are divided into cropped patches of size $1024 \times1024$ pixels with a stride of $824$ pixels, including random flipping is applied as a form of data augmentation. We submit our test set results to the evaluation website.

\noindent{\textbf{DIOR-R}} \cite{dior-r} is an extended version of the DIOR dataset \cite{dior} that shares the same 23,463 optimal remote sensing images but is labeled with oriented bounding boxes. The images are $800\times 800$ pixels and vary in spatial resolution. Each object in the image is categorized into one of $20$ classes. The training procedure for DIOR-R follows the same one implemented on the DOTA-v1.0 dataset.

\noindent{\textbf{HRSC2016}} \cite{hrsc} dataset is a collection of 1061 high-resolution aerial images with oriented bounding box annotations for ship detection. The image sizes range from $300\times 300$ to $1500\times 900$. This dataset encompasses over $25$ categories of ships with a large diversity of shapes, scales, ratios, and orientations annotated. During the training process, the images are resized to $800\times 512$ pixels while preserving their original aspect ratio. Additionally, random flipping is applied to augment the training data.

\input{table/TAB_result_dota}
\input{table/TAB_result_hrsc}
\input{table/TAB_result_diorr}

\input{table/TAB_result_anchor_generation}

\subsection{Implementation details}
We utilize two backbone architectures: ResNet50+FPN and ResNet101+FPN, as well as ResNext101-FPN-DCNv2, all pretrained on ImageNet \cite{imagenet}. Our optimization setup includes the SGD optimizer with an initial learning rate of 0.005 and a decay factor of 10 for each decay step. The weights $w^{AF}$ and $w^{AB}$ are both set to 7.0. We set the momentum to 0.9 and the weight decay to 0.0001. The anchor's scale and ratio is set to 4.0 and 1.0, respectively. For the anchor assigner strategy, we employ the Ratio-based strategy (positive: 0.3; negative: 0.1) \cite{guidedanchoring} for the Anchor-Free head and the Max-IoU strategy (positive: 0.7; negative: 0.3) for the Anchor-based head. Both assigning strategies are followed by random balanced sampling of 256 positive and 256 negative samples. We train HA-RDet for 12 epochs on the DOTA and DIOR-R datasets and 36 epochs on the HRSC2016 dataset. Training is conducted using a single A5000 GPU with a batch size of 2, while inference is performed on a single A5000 GPU.


\subsection{Comparisons with State-of-the-Arts}

The experimental results in Tab.~\ref{tab:result_dota} highlight the performance of HA-RDet on the DOTA-v1 dataset in comparison to prior detectors. While recent methods leveraging ViT-based backbones \cite{wang2022advancing, Li_2023_ICCV, yu2024spatial}, end-to-end training paradigms \cite{zhang2023efficient} and different encoding-decoding \cite{yu2023phase} mechanisms have achieved superior performance—surpassing both traditional one-stage and two-stage detectors as well as our HA-RDet—this study focuses on fair comparisons under the same configuration (Oriented-RCNN, RoI Transformer, S$^2$ANet). \textbf{Specifically, we evaluate HA-RDet against methods using the same backbone architecture and data augmentation strategies.} Future work on these adjustments would be considered.

Under these settings, HA-RDet demonstrates prominent performance, achieving 75.41 mAP and 76.02 mAP with ResNet-50 and ResNet-101 backbones, respectively. The highest accuracy of 77.01 mAP is attained using ResNeXt101-DCNv2. HA-RDet surpasses most anchor-free methods and delivers results competitive with the well-established two-stage Oriented R-CNN.
The incorporation of Orientation-Aware Convolution enables HA-RDet to better align features with object geometry, leading to significantly improved detection accuracy for irregularly shaped objects—such as bridges, storage tanks, and harbors—effectively addressing the challenges posed by high aspect ratios.

On the HRSC2016 dataset, as shown in Tab. \ref{tab:result_hrsc}, HA-RDet achieves competitive results of 90.20 mAP and 95.32 mAP on the VOC2007 and VOC2012 metrics, respectively. These results outperform other training methods that use multiple anchors and demonstrate competitiveness against the $S^2$A-Net, which follows the same training scheme of only using one anchor per location. 



Tab. \ref{tab:result_dior} presents the results obtained from the DIOR-R dataset, where our novel Hybrid-Anchor training schemes approach demonstrates superior performance. Using the backbone ResNet50, our method achieves the highest accuracy at 65.3 mAP, surpassing the well-known two-stage Oriented R-CNN by 0.2 mAP, the one-stage $S^2$A-Net by 3.6 mAP, and the Anchor-free Oriented Proposal Generator (AOPG) by 0.9 mAP.


We conduct further experiments to compare our proposed Hybrid-Anchor training scheme with the well-known anchor-free $S^2$A-Net and anchor-based Oriented R-CNN methods in Tab. \ref{tab:result_compare_s2anet_orientedrcnn}. We evaluate these methods beyond the mean Average Precision (mAP) measure, such as inference speed (FPS), computational resources (VRAM) for storing as well as processing the intermediate data during the training process, and the number of model parameters. The $S^2$A-Net reached the detection result of 74.19 mAP with a fast inference speed of 15.5 FPS; it consumed 4.6 GB of VRAM due to its relatively low number of parameters. Meanwhile, Oriented R-CNN, with 20 anchors in the training stage, attained a slightly better accuracy than $S^2$A-Net but with a trade-off regarding inference speed (13.5 FPS) and higher computational resource requirements (14.2 GB of VRAM). Our HA-RDet follows the Hybrid-Anchor training approach and generates only one anchor per location, similar to $S^2$A-Net. It achieved better detection accuracy (75.41 mAP) while maintaining a stable inference speed of 14.9 FPS. Compared to Oriented R-CNN, our model has a higher parameter requirement for processing but consumes approximately half the amount of VRAM (6.9 GB) due to the significant reduction in the number of generated anchors. Oriented R-CNN utilized dense generated anchors and a complex encoding strategy; therefore, its accuracy and inference speed is slightly higher than our model, but it comes at the cost of significantly higher computational resource requirements. Some visualization of HA-RDet are demonstrated in Fig. \ref{fig:resultdota}

\input{table/TAB_s2anet_orientedrcnn_hardet}
\input{table/TAB_result_on_various_conv}

\subsection{Ablation studies}

\textbf{Effectiveness of O-AwareConv.} As mentioned in section \ref{section_hybrid_anchor_rpn}, we introduced an approach called O-AwareConv that enhances the awareness of ground-truth orientation. Tab. \ref{tab:result_various_conv} compares O-AwareConv against alternative methods under the same configurations to demonstrate its effectiveness in oriented object detection. Our HA-RDet employs Standard Convolution and reaches only 74.1 mAP because the fixed receptive field fails to extract orientation-sensitive features. On the other hand, Deformable Convolution addresses feature misalignment, yet its performance degrades when detecting densely packed and arbitrarily oriented objects in aerial images. O-AwareConv significantly enhances the performance of our proposed detector, surpassing Deformable Convolution by 0.91 mAP and outperforming standard Convolution by 1.31 mAP.

\noindent{\textbf{Effectiveness of Hybrid-Anchor mechanism.}} To evaluate the effectiveness of the Anchor-free head, Anchor-based head, and O-AwareConv in our RPN, we experiment with different HA-RDet settings where omitting certain branches, shown in Tab. \ref{tab:kip_components}. When we excluded the Anchor-free head and relied solely on the anchor-based mechanism, the accuracy slightly decreased to 73.153 mAP compared to using the entire component on DOTA-v1. The role of the Anchor-free head is to provide initial proposals not constrained by predefined anchor boxes. By removing this module, we argue that the detector loses its adaptability to various object sizes and aspect ratios, which can lead to a decrease in performance.

\input{table/TAB_skip_component}

On the other hand, taking out the Anchor-based head indicates a terrible accuracy of 0.038 mAP. The Anchor-based head is responsible for refining the sparsely initial proposals and generating high-quality region proposals. Without this component, the detector lacks the refinement process, leading to imprecise and less reliable region proposals. Furthermore, without the Anchor-based head, the detector heavily relies on the initial proposals generated by the Anchor-free head, which may not be as effective in handling objects with different scales, aspect ratios, or complex backgrounds.

These configurations neglect the use of O-Aware Convolution, which bridges the Anchor-free and Anchor-based heads in our Hybrid-Anchor RPN. By disregarding one of these modules, the detector loses the refinement process,  accurately predicting bounding box regression, and the ability to handle object shape and orientation information, resulting in a significant decline in performance.

\noindent{\textbf{Study on anchor scale.}} We use one preset anchor with one scale and ratio for each location on a feature map following the anchor-free mechanism Tab. \ref{tab:result_anchorgen}. Then, we extend the training longer, and the diminishing of samples happens in models having anchor scale larger than 6 due to the overwhelming number of small-sized instances of the datasets. Finally, we decided to choose the anchor scale of 4 with ratio of 1.0 as the representative for our HA-RDet baseline given its stability, easy convergence, and suitability.

%% file: table/TAB_result_dota.tex
\begin{table*}
  \centering
  \renewcommand\arraystretch{1.2}
  \resizebox{\linewidth}{!}{%
  \begin{tabular}{c|c|c|c|c|c|c|c|c|c|c|c|c|c|c|c|c|c}
    \toprule
    \textbf{Model} & Backbones & PL & BD & BR & GTF & SV & LV & SH & TC & BC &ST & SBF & RA & HA & SP & HC & mAP \\
    \midrule
    \multicolumn{18}{c}{\textit{\textbf{One-stage}}} \\
    \midrule

    {\text{RSDet} \cite{rsdet}}  & ResNet50+FPN
    & 89.3 & 82.7 & 47.7 & 63.9 & 66.8 & 62 & 67.3 & 90.8 & 85.3 & 82.4 & 62.3 & 62.4 & 65.7 & 68.6 & 64.6 & 70.8 \\
    

    {\text{SASM} \cite{hou2022shape}} & ResNet50+FPN &86.42 &78.97&52.47&69.84&77.30&75.99& 86.72 & 90.89 & 82.63 & 85.66 & 60.13 & 68.25 & 73.98 & 72.22 & 62.37 &74.92 \\ 

    {\text{PSCD} \cite{yu2023phase}} & ResNet50+FPN & 89.32 & 82.29 &  37.92 & 71.52 & 78.40 & 66.33 & 78.01 & 90.89 & 84.21 & 80.63 & 60.22 & 64.73 & 59.69  & 68.37 & 53.85 & 71.09 \\
    
    {\text{$S^2$A-Net} \cite{$S^2$A-Net}}  & ResNet50+FPN
    & 89.3 & 80.49 & 50.42 & 73.23 & 78.42 & 77.4 & 86.8 & 90.89 & 85.66 & 84.24 & 62.16 & 65.93 & 66.66 & 67.76 & 53.56 & 74.19 \\

    
    \midrule
    \multicolumn{18}{c}{\textit{\textbf{Two-stage}}} \\
    \midrule
    {\text{RoI Transformer} \cite{roitrans}}  & ResNet101+FPN 
    & 88.64 & 78.52 & 43.44 & 75.92 & 68.81 & 73.68 & 83.59 & 90.74 & 77.27 & 81.46 & 58.39 & 53.54 & 62.83 & 58.93 & 47.67 & 69.56 \\
    
    {\text{ARC} \cite{pu2023adaptive}} & ARC-ResNet50+FPN 
    & 89.40  & 82.48 & 55.33&73.88& 79.37& 84.05& 88.06& 90.90& 86.44& 84.83& 63.63& 70.32& 74.29& 71.91& 65.43& 77.35 \\
    
    {\text{Oriented R-CNN} \cite{orientedrcnn}}  & ResNet50+FPN
    & 89.35 & 81.41 & 52.6 & 75.02 & 79.03 & 82.41 & 87.82 & 90.9 & 86.4 & 85.3 & 63.36 & 65.7 & 68.28 & 70.48 & 57.23 & 75.69 \\

    
    
    {\text{ReDet} \cite{redet}}  & ReResNet50+ReFPN
    & 88.79 & 82.64 & 53.97 & 74.00 & 78.13 & 84.06 & 88.04 & 90.89 & 87.78 & 85.75 & 61.76 & 60.39 & 75.96 & 68.07 & 63.59 & 76.25 \\
    \midrule
    \multicolumn{18}{c}{\textit{\textbf{Anchor-free}}} \\
    \midrule
    {\text{Rotated RepPoints} \cite{reppoint}} & ResNet50+FPN 
    & 83.36 & 63.71 & 36.27 & 51.58 & 71.06 & 50.35 & 72.42 & 90.1 & 70.22 & 81.98 & 47.46 & 59.5 & 50.65 & 55.51 & 53.07 & 59.15 \\
    
    
    

    {\text{CFA} \cite{cfa}}  & ResNet101+FPN
    & 89.26 & 81.72 & 51.81 & 67.17 & 79.99 & 78.25 & 84.46 & 90.77 & 83.4 & 85.54 & 54.86 & 67.75 & 73.04 & 70.24 & 64.96 & 75.05 \\

    {\text{AOPG} \cite{dior-r}}  & ResNet50+FPN   & 89.27 & 83.49 & 52.50 & 69.97 & 73.51 & 82.31 & 87.95 & 90.89 & 87.64 & 84.71 & 60.01 & 66.12 & 74.19 & 68.30 & 57.80 & 75.24 \\

    \midrule
    \multicolumn{18}{c}{\textit{\textbf{Ours}}} \\
    \midrule
    {\textbf{HA-RDet}} & ResNet50+FPN 
    & 88.98 & 83.34 & 55.02 & 72.04 & 78.25 & 81.43 & 86.94 & 90.91 & 85.46 & 85.48 & 61.33 & 61.57 & 74.50 & 68.97 & 56.86 & 75.408 \\  
    
    {\textbf{HA-RDet}}  & ResNet101+FPN 
    & 89.12 & 84.95 & 54.81 & 69.33 & 77.67 & 82.91 & 87.54 & 90.87 & 86.06 & 86.03 & 64.19 & 61.90 & 74.25 & 70.99 & 65.43 & 76.02 \\
    
    {\textbf{HA-RDet}}  & ResNeXt101\_DCNv2+FPN 
    & 89.31 & 85.01 & 57.03 & 68.94 & 78.14 & 83.35 & 87.61 & 90.91 & 87.32 & 86.67 & 65.14 & 63.05 &76.35& 72.29& 63.98& 77.012 \\
    \bottomrule
  \end{tabular}}
  \caption{Comparison of SOTA results on the oriented object detection DOTA 1.0 task. We report the results of methods using the same training configuration on DOTA 1.0}
  \label{tab:result_dota}
\end{table*}

%% file: table/TAB_result_hrsc.tex

\begin{table*}
  \centering
  \renewcommand\arraystretch{1.1}
  \resizebox{\linewidth}{!}{%
  \begin{tabular}{c|c|c|c|c|c|c|c|c|c}
    \toprule
    \textbf{Model} 
    & \multicolumn{1}{l}{R3Det \cite{r3det}} 
    & \multicolumn{1}{l}{AO2-DETR \cite{aood}} 
    & \multicolumn{1}{l}{Gliding Vertex \cite{glidingvertex}} 
    & \multicolumn{1}{l}{$S^2$A-Net \cite{$S^2$A-Net}}
    & \multicolumn{1}{l}{ReDet \cite{redet}}
    & \multicolumn{1}{l}{AOPG \cite{dior-r}}
    & \multicolumn{1}{l}{Oriented RepPoints \cite{orientedreppoint}} 
    & Oriented R-CNN \cite{orientedrcnn}
    & \textbf{HA-RDet} \\
    \midrule
    \textbf{\#Anchors} 
        & \multicolumn{1}{c}{21} 
        & \multicolumn{1}{c}{-} 
        & \multicolumn{1}{c}{20} 
        & \multicolumn{1}{c}{1} 
        & \multicolumn{1}{c}{20}
        & \multicolumn{1}{c}{1} 
        & \multicolumn{1}{c}{9} 
        & 20
        & 1 \\
    \midrule
    {\textbf{mAP}} 
        & \multicolumn{1}{c}{87.14}
        & \multicolumn{1}{c}{88.12}
        & \multicolumn{1}{c}{88.20}
        & \multicolumn{1}{c}{90.14} 
        & \multicolumn{1}{c}{90.2} 
        & \multicolumn{1}{c}{90.34} 
        & \multicolumn{1}{c}{90.38} 
        & 90.4
        & 90.20 / \textbf{95.32}$^*$ \\
    \bottomrule
  \end{tabular}}
  \caption{Results on HRSC2016 with \#Anchors represents the quantity of anchors generated per location on a feature map.}
  \label{tab:result_hrsc}
\end{table*}

%% file: table/TAB_result_diorr.tex

\begin{table*}
  \centering
  \renewcommand\arraystretch{1.1}
  \resizebox{16.5cm}{!}{%
  \begin{tabular}{c|c|c|c|c|c|c|c|c}
    \toprule
    \textbf{Model} 
    & \multicolumn{1}{l}{R3Det \cite{r3det}} 
    & \multicolumn{1}{l}{CFA \cite{cfa}} 
    & \multicolumn{1}{l}{$S^2$A-Net \cite{$S^2$A-Net}}  
    & \multicolumn{1}{l}{Rotated FCOS \cite{fcos}} 
    & \multicolumn{1}{l}{RoI Transformer \cite{roitrans}}
    & \multicolumn{1}{l}{AOPG \cite{dior-r}}
    & Oriented R-CNN \cite{orientedrcnn}
    & \textbf{HA-RDet} \\
    \midrule
    \textbf{mAP}
    & \multicolumn{1}{c}{57.2} 
    & \multicolumn{1}{c}{60.8} 
    & \multicolumn{1}{c}{61.7} 
    & \multicolumn{1}{c}{62.4} 
    & \multicolumn{1}{c}{63.9} 
    & \multicolumn{1}{c}{64.4} 
    & 65.1
    & \textbf{65.3} \\
    \bottomrule
  \end{tabular}}
  \caption{Experimental comparison of SOTA methods on DIOR-R, trained using \textbf{ResNet50+FPN}}
  \label{tab:result_dior}
\end{table*}

%% file: table/TAB_result_anchor_generation.tex

\begin{table*}
  \centering
  \renewcommand\arraystretch{1.1}
  \resizebox{\linewidth}{!}{%
  \begin{tabular}{c|c|c c c c c c c c c c c c c c c|c}
    \toprule
    \textbf{Model (anchor scale)} & Backbones & PL & BD & BR & GTF & SV & LV & SH & TC & BC &ST & SBF & RA & HA & SP & HC & mAP \\
    \midrule
    {\textbf{HA-RDet (4)}} & ResNet50 + FPN
        & 88.98 & \textbf{83.34} & \textbf{55.02} & 72.04 & \textbf{78.25} & \textbf{81.43} & \textbf{86.94} & \textbf{90.91} & 85.46 & 85.48 & 61.33 & 61.57 & 74.50 & \textbf{68.97} & 56.86 & \textbf{75.408} \\
    \midrule
    {\textbf{HA-RDet (6)}} & ResNet50 + FPN
        & \textbf{89.01} & 81.84 & 54.34 & \textbf{73.72} & 72.59 & 77.52 & 86.88 & 90.9 & \textbf{87.66} & \textbf{85.75} & \textbf{61.55} & \textbf{61.91} & 75.69 & 66.54 & \textbf{59.72} & 75.032 \\
    \midrule
    {\textbf{HA-RDet (8)}} & ResNet50 + FPN
        & 88.97 & 80.79 & 53.18 & 70.64 & 70.96 & 76.49 & 78.98 & 90.9 & 85.57 & 84.58 & 59.29 & 60.83 & \textbf{75.91} & 66.49 & 53.28 & 73.122 \\
    \bottomrule
  \end{tabular}}
  \caption{Models with anchor scales above 6 perform better on large objects, while smaller scales yield higher accuracy for small instances}
  \label{tab:result_anchorgen}
\end{table*}

%% file: table/TAB_s2anet_orientedrcnn_hardet.tex
\begin{table}
  \centering
  \renewcommand\arraystretch{1.2}
  \resizebox{8.3cm}{!}{%
  \begin{tabular}{c|c|c|c}
    \toprule
    \textbf{} & \textbf{$S^2$A-Net \cite{$S^2$A-Net}} & \textbf{Oriented R-CNN \cite{orientedrcnn}} & \textbf{HA-RDet (Ours)} \\
    \midrule
    {\text{\textbf{\#Anchors}}} & 1
    & 20 & 1  \\
    {\text{\textbf{mAP}}}  & 74.19
    & 75.69 & 75.41  \\
    {\text{\textbf{FPS}}}  & 15.5
    & 13.5 & 14.9  \\
    {\text{\textbf{VRAM (GB)}}}  & 4.6
    & 14.2 & 6.9  \\
    {\text{\textbf{\#params}}}  & 38,599,976
    & 41,139,754 & 55,715,449  \\
    \bottomrule
    \end{tabular}}
  \caption{Evaluation results of HA-RDet with $S^2$A-Net and Oriented R-CNN on the \textbf{ResNet50+FPN} on DOTA. Anchors are saved under CUDA tensors for fast computation}
  \label{tab:result_compare_s2anet_orientedrcnn}
\end{table}

%% file: table/TAB_result_on_various_conv.tex
\begin{table}
  \centering
  \renewcommand\arraystretch{1.2}
  \resizebox{7.0cm}{!}{%
  \begin{tabular}{c|c}
    \toprule
    \textbf{Convolution method} & \textbf{mAP} \\
    \midrule
    Standard Convolution & 74.1 \\
    Deformable Convolution \cite{dcn} & 74.5 (\(\uparrow\) 0.4) \\
    Orientation-aware Convolution (Ours) & \textbf{\textcolor{black}{75.41}} (\(\uparrow\) 1.3) \\
    \bottomrule
  \end{tabular}}
  \caption{Comparison of Orientation-aware Convolution accuracy with other convolutions on the DOTA-v1 dataset.}
  \label{tab:result_various_conv}
\end{table}

%% file: table/TAB_skip_component.tex
\begin{table}
  \centering
  \renewcommand\arraystretch{1.2}
  \resizebox{8.3cm}{!}{%
  \begin{tabular}{p{3.5cm}|ccc}
    \toprule
    \textbf{Module} & & \textbf{Our Hybrid-Anchor RPN} & \\
    \midrule
    Anchor-free head & \checkmark & \checkmark & - \\
    Anchor-based head & \checkmark & - & \checkmark \\
    O-AwareConv & \checkmark & - & - \\
    \midrule
    mAP & \textcolor{black}{\textbf{75.408}} & 0.038 & 73.153  \\
    \bottomrule
  \end{tabular}}
  \caption{Experimental evaluation of our Hybrid-Anchor RPN by taking out a couple of modules, using ResNet50+FPN on DOTA.}
  \label{tab:kip_components}
\end{table}

%% file: sec/5_conclusion.tex
\section{Discussion}
We propose HA-RDet, a Hybrid-Anchor training scheme that integrates the novel O-AwareConv to tackle two critical challenges in oriented object detection: the scarcity of positive samples and the risk of over-encoding geometric information. HA-RDet bridges the gap between one-stage and two-stage detection paradigms, striking a balance between training/inference efficiency and detection accuracy.
Our method is particularly well-suited for Earth Observation applications, where aerial and satellite imagery often contain densely packed, arbitrarily oriented objects. Use cases include maritime monitoring, urban development assessment, and disaster response. The ability to detect rotated objects precisely is crucial in these contexts, where misalignment can lead to misclassification or missed detections.
HA-RDet’s computational efficiency, achieved via a reduced anchor set and lightweight orientation-aware refinement, makes it deployable on edge devices such as drones or low-power remote sensing platforms. Notably, HA-RDet has been quantized and deployed on campus drone systems, enabling continuous monitoring and autonomous governance.
